\documentclass{ecai} 

\usepackage{latexsym}
\usepackage{amssymb}
\usepackage{amsmath}
\usepackage{amsthm}
\usepackage{booktabs}
\usepackage{enumitem}
\usepackage{graphicx}
\usepackage{color}
\usepackage{multirow}

\usepackage[T1]{fontenc}

\newcommand{\main}{f} 

\newcommand{\nna}{f_{A}} 
\newcommand{\nnb}{f_{B}} 
\newcommand{\nnc}{f_{C}} 
\newcommand{\nngtsrb}{f_{GTSRB}}
\newcommand{\nncub}{f_{CUB}}
\newcommand{\nnplaces}{f_{Places}}
\newcommand{\nnfashion}{f_{Fashion}}
\newcommand{\nnimagenet}{f_{ImageNet}}

\newcommand{\concept}[1]{\mathsf{#1}} 

\newcommand{\probe}{g} 

\newcommand{\datas}{\mathcal{D}}

\newcommand{\picscale}{0.95}

\begin{document}


\begin{frontmatter}

\title{On the Performance of Concept Probing:\\ The Influence of the Data \\
{\LARGE -- Extended Version --\footnotemark[1]}}

\author{\fnms{Manuel}~\snm{de Sousa Ribeiro}}
\author{\fnms{Afonso}~\snm{Leote}}
\author{\fnms{Jo{\~{a}}o}~\snm{Leite}}
 
\address{NOVA LINCS, NOVA School of Science and Technology, NOVA University Lisbon, Portugal\\ \{mad.ribeiro, a.leote, jleite\}@fct.unl.pt}


\begin{abstract}
Concept probing has recently garnered increasing interest as a way to help interpret artificial neural networks, dealing both with their typically large size and their subsymbolic nature, which ultimately renders them unfeasible for direct human interpretation. 
Concept probing works by training additional classifiers to map the internal representations of a model into human-defined concepts of interest, thus allowing humans to peek inside artificial neural networks. 
Research on concept probing has mainly focused on the model being probed or the probing model itself, paying limited attention to the data required to train such probing models. 
In this paper, we address this gap. Focusing on concept probing in the context of image classification tasks, we investigate the effect of the data used to train probing models on their performance. 
We also make available concept labels for two widely used datasets.
\end{abstract}

\end{frontmatter}


\footnotetext[1]{This is an extended version of \citep{deSousaRibeiro2025ECAI}.}
\addtocounter{footnote}{+1}


\section{Introduction} \label{Sec:introduction}

In this paper, we investigate how different characteristics of the data required for concept probing in neural network models - such as size, number of features, provenance, and quality -  affect the performance of popular concept probing methods.

The application of neural networks in tasks that may influence the health or safety of individuals highlights the urgency for methods to interpret neural network models and validate how they achieve their results \cite{Wang2022}.
The field of Explainable AI seeks to address this issue, providing methods to help improve a model's interpretability.

One of the most prominent methodologies for interpreting neural network models is known as \emph{concept probing} \cite{Alain2017}. 
Concept probing consists of training a model -- referred to as \textit{probe} -- to identify a given concept of interest from the activations of an existing neural network model. In this way, one is able to estimate the mutual information between the representations of a model and some concept of interest \cite{Pimentel2020}, with higher prediction accuracy implying that the representations carry more information about the concept of interest.
Concept probing provides some insight into what is encoded in the representations of a neural network model without the need to retrain or modify the model being examined. It allows humans to define and experiment with their own concepts of interest according to their needs and requirements.

Apart from providing humans with a way to \textit{peek} into what is encoded within a neural network model, the trained probes can be used to map subsymbolic internal representations of a model into symbolic observations regarding a model's internal state. These probes can then be leveraged by interpretability and explainability methods, e.g., to test the behavior of a model's internal representations \cite{Lovering2022}, to produce human-understandable justifications for a neural network's output \cite{deSousaRibeiro2021,Ferreira2022}, or to generate and study a model's counterfactual behavior \cite{Tucker2021}, to name a few.

This methodology has sparked interest in different communities, which have mainly focused on matters regarding the design of the probing models and the interpretation of their results. However, as stressed in \cite{Belinkov2022}, matters regarding the data that is necessary to properly develop concept probes have been largely overlooked.
In the absence of such studies, practitioners have to rely on their intuitions, which often leads to the widespread of unproven \emph{folk theories} concerning crucial issues on concept probing.
Questions such as ``How much data does concept probing require?'', ``Can concept probing handle large models?'', ``Does one need a \emph{fresh} dataset for concept probing? Or can one repurpose the data used in the development of the probed model?'', and ``Does concept probing require high-quality data?'' have long been asked in the community, with their answers often taken for granted without dedicated studies supporting them.
This has significant repercussions on the research being carried out. E.g., \cite{Palsson2024} describes using vast amounts of data for probing because the number of required samples is unclear.
This is a significant issue as data is generally the major cost of employing concept probing, since labeled data is necessary to probe for each concept of interest. 
It is thus essential to have a clearer understanding of how the data used to train  probing models affects their performance, as it will allow practitioners to better assess when it is feasible to apply concept probing and how to do it  effectively -- understanding the existing trade-offs and avoiding common pitfalls, like using an inadequate amount of training samples or neglecting to assess their data's quality. 

In this paper, we focus on some of the most common yet scarcely addressed questions about how the characteristics of the datasets used to train such probing models might affect their performance.
Based on a representative selection of image classification datasets, neural network models, and probing architectures from the literature, we perform a thorough experimental evaluation of the influence of data in concept probing along four main dimensions, namely: i) testing the effect of the probing models' train data size on their performance; ii) inspecting how the size of the probed model impacts the probing model's performance; iii) examining the impact of reusing data from the probed model train set to train the probing models; iv) verifying how the quality of the probing models' training data affects their performance. While doing so, we also examine the behavior of the most popular probing architectures on various image classification neural network models and datasets.

Our results confirm some existing ideas on the influence of data in concept probing but also support some surprising conclusions related, e.g., to data reuse and the robustness of probing models, that ultimately impact probing's applicability.

The remainder of this paper is organized as follows: Section \ref{Sec:concept_probing} offers an overview of concept probing, followed by a description of the methods and experimental setup in Section \ref{Sec:experimental_setup}. 
Then, in Section \ref{Sec:dataset_size} through \ref{Sec:data_quality}, we present and discuss our experimental results. 
Finally, we discuss some related work in Section \ref{Sec:related_work} and conclude with a summary of our findings in Section \ref{Sec:conclusions}.


\section{Concept Probing} \label{Sec:concept_probing}

The intuition behind concept probing is that neural networks distill useful representations over the course of their layers, which allow them to gradually abstract away from the input space and can finally be used to produce their expected outputs.
Thus, one could monitor such representations by observing the activations of a model's units and assess how well they represent some given concept of interest -- also referred to as \emph{property} in the literature \cite{Belinkov2022}.

Hence, in concept probing, a given neural network model $\main\colon x \mapsto y$ -- which we refer to as \emph{original model} -- is probed for representations with information regarding some human-defined concept of interest $\concept{C}$.
The way the probing occurs is by mapping the internal representations of $\main$ to $\concept{C}$ by means of another model $\probe$ -- referred to as \emph{probe}.
Typically, probes do not map all internal representations of some original model $\main$, but a subset of those. 
E.g., the activations produced by all units of a given layer of the model $\main$.
Let $\main_{u}(x)$ denote the representation of $x$ at some sequence of units $u$ from model $\main$, i.e., the activations generated by model $f$ when fed with input $x$ at units $u$.
A probing model $\probe\colon \main_{u}(x) \mapsto c$ is a model trained to identify some concept of interest $\concept{C}$, with $c$ representing its value, from some internal representations $u$ of the original model $\main$.
As an example, $\main$ might be a convolutional neural network trained to classify a bird's species from an image, and $\probe$ might be a classifier mapping the output of the convolutional part of $\main$ to some concept of interest $\concept{C}$, such as whether a bird's \emph{yellow belly} is visible in the image -- a concept which is relevant to identify some bird species.

The dataset to train a probe model $\probe$ is different from the one used to train $\main$, $\datas_\main = \{x^{(i)}, y^{(i)}\}$, which is composed of input samples $x$ of $\main$ and their respective outputs.
The probe's dataset should be composed of the observed activations of $\main$ for a set of instances, along with annotations for the corresponding value of the concept of interest $\concept{C}$ for that instance, i.e., $\datas_\probe = \{\main_{u}(x^{(i)}), c^{(i)}\}$. This dataset is generally balanced wrt. the concept values ($c^{(i)}$).
The semantics of concept of interest $\concept{C}$ is given extensionally by the dataset $\datas_\probe$.

The performance of $\probe$ is measured after its training on a separate test dataset $\datas'_\probe$, similar to $\datas_\probe$ but made of fresh instances. Since the data is typically balanced, the accuracy of $\probe$ on $\datas'_\probe$ is generally considered, with higher accuracies indicating that the units $u$ probed by $\probe$ carry more information for representing concept $\concept{C}$.
Throughout our experiments, we discuss how different characteristics of the dataset $\datas_\probe$ might affect the performance of a probing model $\probe$.

The particular architecture of a probing model is still a debated issue, with some works advising the use of simpler linear probing models \cite{Alain2017,Liu2019}, while others suggest that more complex probing architectures should be considered \cite{Pimentel2020a,Pimentel2020,Tucker2021}. In this work, we consider both types of probing architectures.


\section{Methods and Experimental Setup}
\label{Sec:experimental_setup}

To study how various characteristics of the data used for concept probing influence the resulting probes, one should consider various datasets, original models, and probing architectures. 
In this section, we describe the probing models' architectures used in our empirical study. Subsequently, we introduce the datasets used in the experiments and their respective original models.

\subsection{Probing Architectures}

We consider five different architectures for probe $\probe$ -- a linear logistic regression classifier (Logistic); a linear ridge classifier (Ridge); a LightGBM decision tree (LGBM) \cite{Ke2017}; a neural network (NN); and a mapping network probe (MapNN) \cite{deSousaRibeiro2021} -- ensuring that our results are not too tied to a particular kind of probe.
When training all models, $25\%$ of the training data was used as a validation set. If early stopping is used, we consider a patience value of $15$.
For the linear Ridge probe, we perform a hyperparameter search over the alpha values of $[0.01$, $0.05$, $0.1$, $0.5$, $1$, $5$, $10$, $50$, $100]$.
The LGBM probe is used with default parameters, except for the minimal number of instances at a terminal node, which is reduced when the default value is too large for the model to train. The validation set is used together with early stopping to select the number of boosting rounds.
The NN probe has a feedforward architecture with ReLU non-linearity and a hidden layer of size 10.
The MapNN probe shares the same architecture as the NN, but L1 regularization is applied to its weights with a strength of $0.001$. Mapping network probes are trained using the input reduce procedure described in \cite{deSousaRibeiro2021}, a procedure that iterates a set of layers of a model to identify a subset of units from which some concept of interest might be probed from, with a patience value of $3$ and the ranking of each feature being given by its maximum absolute weight.
Early stopping is used to select the number of training epochs for both the neural network and the mapping network probes.

All probes are trained using the activations produced by some given layer of an original model $\main$, with the exception of the mapping network probes, which use the input reduce procedure, and thus might select units $u$ from various layers of the original model $\main$.


\begin{figure}[tb]
	\centering
	
	\includegraphics[width=\picscale\linewidth]{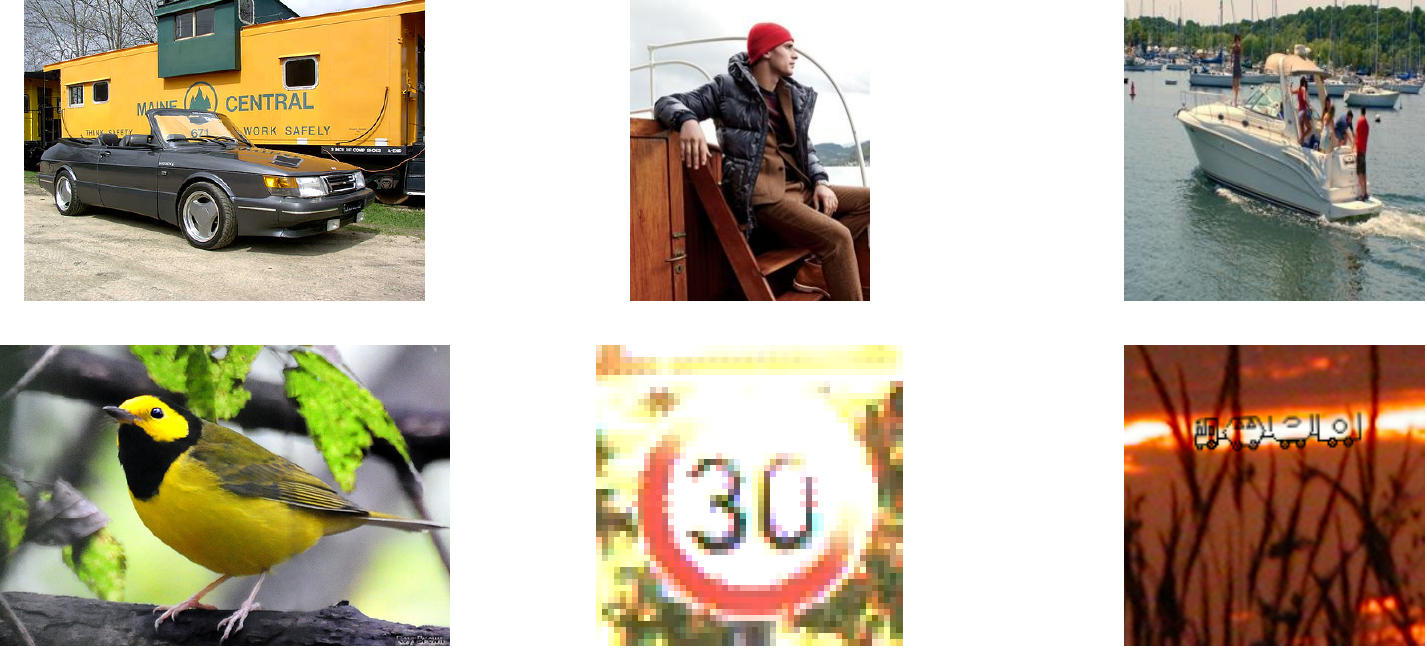}

	\caption{Sample images from each dataset: ImageNet \cite{Deng2009}, DeepFashion \cite{Liu2016}, Places365 \cite{Zhou2018}, CUB \cite{Wah2011}, GTSRB \cite{Stallkamp2011}, and XTRAINS \cite{deSousaRibeiro2020}.}
	\label{fig:sample_images}
\end{figure}

\subsection{Datasets and Original Models}

\begin{table*}[t]
{\center
{\scriptsize
\begin{tabular}{@{}rcccccc@{}}
\toprule
Property               & ImageNet              & DeepFashion               & Places365                & CUB                   & GTSRB                 & XTRAINS{$_\concept{A;B;C}$}               \\ \midrule
Type                   & Real                  & Real                  & Real                  & Real                  & Real                  & Synthetic             \\
Task                   & Multiclass            & Multiclass            & Multiclass            & Multiclass            & Multiclass            & Binary                \\
\# Class               & $1000$                  & $12$                    & $365$                   & $200$                   & $43$                    & $1$; $1$; $1$               \\
Image Entropy          & $0.91$                  & $0.74$                  & $0.93$                  & $0.89$                  & $0.79$                  & $0.76$                  \\
Compression Ratio       & $1.09$                  & $1.05$                  & $1.08$                  & $1.11$                  & $2.44$                  & $2.00$                  \\
Concept Separability   & $0.59$                  & $0.58$                  & $0.63$                  & $0.63$                  & $0.84$                  & $0.52$                  \\
Original Model         & ViT                   & ResNet101             & ResNet50              & ResNet50              & MobileNetV2           & VGGNet                \\
Pre-trained            & Yes                   & Yes                   & Yes                   & Yes                   & No                    & No                    \\
Input Shape            & $224\!\times\!224\!\times\!3$ & $224\!\times\!224\!\times\!3$ & $224\!\times\!224\!\times\!3$ & $224\!\times\!224\!\times\!3$ & $128\!\times\!128\!\times\!3$ & $152\!\times\!152\!\times\!3$ \\
Model Size (\# params) & $85.8$ M                & $44.5$ M                & $28.4$ M                & $28.4$ M                & $2.3$ M                 & $421$; $421$; $561$ K       \\ \bottomrule
\end{tabular}
}
\caption{Experimental setting of each dataset.}
\label{tab:experimental}
}
\end{table*} 

We experiment with probing eight trained neural network models, from six image classification datasets.
Table \ref{tab:experimental} presents a summary of the experimental setting for each dataset.
These datasets are representative of a variety of different image classification settings where concept probing might be applied, ranging from binary classification to multiclass classification with $1000$ classes.
The image complexity of the datasets, measured by their \emph{entropy} and \emph{compression ratio} \cite{Yu2013}, reveals a set of settings with moderately and highly complex images.
In some cases, the probed concepts are closer to the images, indicated by a high concept separability score -- measured as the accuracy of a linear model predicting the concepts from the images -- while in others, the concepts are more abstract.
Simultaneously, both simpler and more complex original models are considered, illustrating different circumstances for the application of concept probing.

\paragraph{ImageNet}\!\!\!\!\!\! \cite{Deng2009}: this dataset has labels for $1000$ varied object classes.
As the original model, we consider the Vision Transformer (ViT) from \cite{Wu2020} with $84.2\%$ accuracy on the test set, which we refer to as $\nnimagenet$.
We probe $11$ random concepts from the related ImageNet Object Attributes dataset \cite{Russakovsky2010}, a subset of ImageNet with labels for attributes related to colors, shapes and textures: $\sf\exists has\-Texture.Fur\-ry$, $\sf \exists has\-Color.Green$, $\sf \exists has\-Shape.Rec\-tan\-gu\-lar$, $\sf\exists has\-Color.Red$, $\sf\exists has\-Shape.Round$, $\sf\exists has\-Texture.Shin\-y$, $\sf\exists has\-Pattern.Spot\-ted$, $\sf\exists has\-Pattern.Striped$, $\sf\exists has\-Texture.Wet$, $\sf\exists has\-Texture.Wood\-en$, and $\sf\exists has\-Color.Yel\-low$.

\paragraph{DeepFashion}\!\!\!\!\!\! \cite{Liu2016}: this dataset contains images of $50$ different clothes categories, with $1000$ different labeled clothing-related attributes.
It is known to be highly imbalanced and noisily labeled \cite{Song2024}.
As the original model, we consider a ResNet101 pre-trained in ImageNet \cite{Deng2009} and finetuned to classify the $12$ most populated clothes categories in this dataset, achieving a test accuracy of $59.60\%$ in a balanced test set of $16\ 800$ images. We refer to this model as $\nnfashion$. 
From this dataset, we probe the following $11$ clothing-related attributes: $\sf Body\-con$, $\sf \exists has.FauxLeather$, $\sf \exists has.Graph\-ic$, $\sf \exists has.Hood\-ed$, $\sf Maxi$, $\sf Mi\-di$, $\sf Mo\-to$, $\sf Pen\-cil$, $\sf \exists has.Racer\-back$, $\sf Skater$,    and $\sf \exists has.Strap\-less$.

\paragraph{Places365}\!\!\!\!\!\! \cite{Zhou2018}:
this dataset is composed of images of $365$ scene categories.
As the original model, we consider the ResNet50 from \cite{Zhou2018} trained to classify each scene category with a test accuracy of $54.65\%$, which we refer to as $\nnplaces$. 
As this dataset has no additional annotations, we probe the $10$ concepts defined in the CIFAR-10 dataset \cite{Krizhevsky2009}: $\sf Air\-plane$, $\sf Au\-to\-mo\-bile$, $\sf Bird$, $\sf Cat$,  $\sf Deer$, $\sf Dog$, $\sf Frog$, $\sf Horse$, $\sf Ship$, and $\sf Truck$.

\paragraph{Caltech-UCSD Birds-200-2011 (CUB)}\!\!\!\!\!\! \cite{Wah2011}:
this dataset has images of birds from $200$ species. 
Each image is labeled with various attributes representing visual concepts that are described as relevant to the identification of the bird species.
From these attributes, we randomly selected the following $11$ concepts to probe:
$\sf \exists has\-Bill\-Shape.Nee\-dle$, $\sf \exists has\-Bill\-Shape.Hooked\-Sea\-bird$, $\sf \exists has\-Head\-Pat\-tern.Striped$, $\sf \exists has\-Breast\-Col\-or.Yel\-low$, $\sf \exists has\-Throat\-Col\-or.Red$, $\sf \exists has\-Eye\-Col\-or.Red$, $\sf \exists has\-Bel\-ly\-Col\-or.Blue$, $\sf \exists has\-Bel\-ly\-Col\-or.Yel\-low$, $\sf \exists has\-Shape.Duck\-Li\-ke$, $\sf \exists has\-Crown\-Col\-or.White$, $\sf \exists has\-Crown\-Col\-or.Red$.
As the original model, we consider the ResNet50 from \cite{Taesiri2022} pre-trained in the iNaturalist dataset \cite{Horn2018} and finetuned in the CUB dataset, achieving an accuracy of about $85.83\%$ on the dataset's test data. We refer to this model as $\nncub$.

As observed in \cite{Zhao2019, Koh2020}, CUB's attributes are noisily labeled. We relabeled a subset of the data for each of the probed concepts \footnote{Details are made available in Appendix \ref{Sec:appendix_cub}.}, which we use in our experiments. We further discuss this topic in Section \ref{sec:data_quality}.

\paragraph{German Traffic Sign Recognition Benchmark (GTSRB)}\!\!\!\!\!\! \cite{Stallkamp2011}:
this dataset has images of $43$ types of traffic signs. 
Additionally, we consider an ontology and labels based on the 1968 Convention on Road Signs and Signals \cite{eutrafficconvention}, which describes each type of traffic sign in the dataset based on visual concepts\footnote{Details are made available in Appendix \ref{Sec:appendix_gtsrb}.}.
For example, a stop sign is described as having an octagonal shape, a red ground color, and a white `stop' symbol.
We probe the following $10$ random concepts defined in the ontology: $\sf\exists has\-Sym\-bol.\top$, $\sf\exists has\-Bar.Black$, $\sf\exists has\-Sym\-bol.Black$, $\sf Blue$, $\sf\exists has\-Sha\-pe.Cir\-cu\-lar$, $\sf Pole$, $\sf\exists has\-Ground.Red$, $\sf\exists has\-Sym\-bol.Speed\-80$,  $\sf\exists has\-Sha\-pe.Tri\-an\-gu\-lar$, and $\sf\exists has\-Ground.Whi\-te$.
As the original model, we trained a MobileNetV2 \cite{Sandler2018} -- which we refer to as $\nngtsrb$ -- to identify the class of the traffic sign in an image, achieving an accuracy of about $98\%$ on the dataset's test data. 

\paragraph{Explainable Abstract Trains Dataset (XTRAINS)}\!\!\!\!\!\! \cite{deSousaRibeiro2020}: 
this synthetic dataset contains representations of trains over landscape backgrounds. It is accompanied by an ontology describing how different concepts in the dataset relate to each other, with each concept having its own visual characteristics.
Three different types of trains ($\concept{TypeA}$, $\concept{TypeB}$, and $\concept{TypeC}$) are defined based on these concepts.
E.g., the concept of $\concept{PassengerCar}$ is visually represented by a wagon having at least one circle inside of it, with the dataset's ontology stating that trains having at least two passenger cars are $\concept{PassangerTrain}$s, and that such trains are $\concept{TypeB}$ trains.
The dataset provides labels for all of these concepts.
As original models, we consider the three VGGNet \cite{Simonyan2015} models from \cite{Ferreira2022} -- referred to as $\nna$, $\nnb$, and $\nnc$ -- trained to identify trains of the corresponding type, each achieving a test accuracy of about $99\%$. 
We probe the $11$ concepts examined in \cite{deSousaRibeiro2021}: $\sf Emp\-ty\-Train$,  $\sf Freight\-Train$, $\sf \exists has.Freight\-Wa\-gon$, $\sf Long\-Train$, $\sf \exists has.Long\-Wagon$, $\sf Mixed\-Train$, $\sf \exists has.Open\-Roof\-Car$, $\sf Pas\-sen\-ger\-Train$, $\sf \exists has.Re\-in\-forced\-Car$, $\sf Ru\-ral\-Train$, and $\sf War\-Train$.

\begin{figure*}[t]
\centering
\includegraphics[width=\picscale\textwidth]{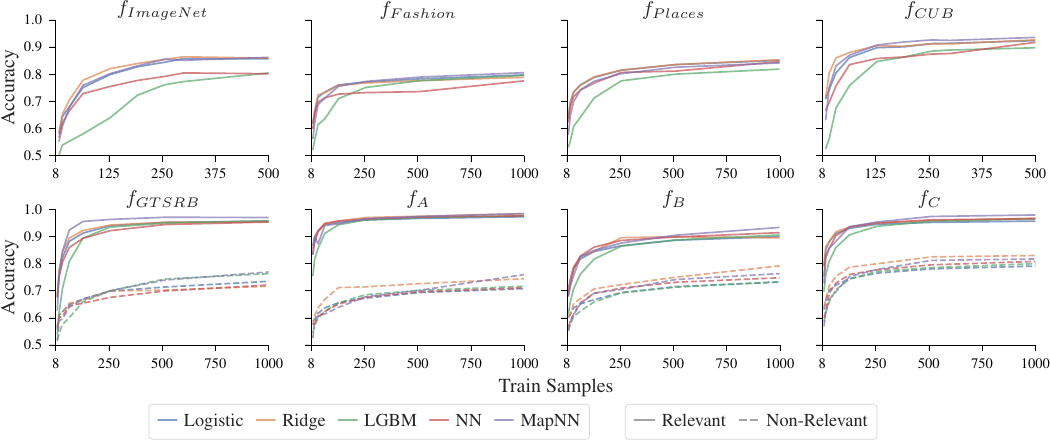}
\caption{Probe performance by amount of training data.}
\label{fig:dataset_size}
\end{figure*}

\paragraph{Probed Activations}

To avoid having our results too tied to the choice of probing the activations of a particular layer of a model, we considered the activations produced by a set of representative layers for each model, taking into account the model's architecture and concepts that were being probed.

For the VGGNet models ($\nna$, $\nnb$, $\nnc$), which have a clear separation between convolutional and dense parts, we probe the activations resulting from their convolutional part and each hidden dense layer.
For the MobileNetV2 model ($\nngtsrb$), which is composed of $17$ residual layers, we consider the activations of every fourth residual layer.
For the ResNet50 ($\nnplaces$, $\nncub$) and ResNet101 ($\nnfashion$) models, which are composed of $4$ residual blocks, we probe the activations produced by the first and last convolution blocks of the fourth residual block, as the activations resulting from the previous blocks seemed to not yet capture the probed concepts.
Similarly, for the ViT model ($\nnimagenet$), we consider the activations resulting from the attention and feedforward layers of the model's last transformer block.

\paragraph{Concept Relevancy}

In our experiments, we distinguish between two sets of probed concepts: relevant and non-relevant. Informally, a concept is said to be relevant to another if there is any circumstance where knowledge relative to the former allows us to infer knowledge about the latter.
E.g., consider $\nncub$, which identifies bird species in CUB. All Hooded warbler birds have a yellow belly; thus, yellow belly is a relevant concept to Hooded warbler birds, as knowing about it might allow us to infer whether a bird is a Hooded warbler. 
\cite{deSousaRibeiro2021} provides a more formal definition of relevancy for concepts defined through ontologies in Description Logics, which we use to determine relevant concepts in the setting of the GTSRB and XTRAINS datasets. 
For the CUB dataset, although no ontology is provided, \cite{Wah2011} states that the attributes were selected based on an expert's guide on identifying birds, and thus all concepts are considered to be relevant.
Similarly, while no ontology is provided for the ImageNet, DeepFashion, and Places365 datasets, we believe the probed concepts are relevant -- at least in the informal sense -- to the task being performed by the respective original models.

While being a relevant concept does not provide guarantees concerning concept probing, since a model might learn to perform its task in a way that does not follow our human ontology of a domain, we nevertheless expect that, on average, probing for relevant concepts should yield better results than probing for non-relevant ones.

\paragraph{Methodological Considerations}

All experimental results in this paper are averaged over $5$ repetitions, using different balanced sets of samples for training, validation, and testing. Due to the variance in the amount of available data in each dataset, we test the trained probes with balanced sets of $1\ 000$ test samples for the Places365 and XTRAINS datasets, $200$ for DeepFashion and GTSRB, and $100$ for the ImageNet and CUB datasets. 
Except for Section \ref{sec:dataset_size}, where the effect of train data size is discussed, all probes were trained using a balanced set of $500$ samples. For some concepts in the ImageNet and CUB datasets, there was insufficient data to produce a balanced set of $500$ samples, so the nearest balanced amount was considered.


\section{Effect of Train Dataset Size}
\label{sec:dataset_size}
\label{Sec:dataset_size}

Perhaps the first and most prevalent question when discussing the feasibility of concept probing is, ``How much data is necessary?''
This is a sensible question to ask, given that for each concept of interest that one aims to probe, a set of labeled data $\datas_\probe$ is required, making data one of the main costs for applying concept probing.

An underlying assumption shared by those working on concept probing is that neural networks distill, throughout their layers, computationally useful representations with information that allows them to address their tasks \cite{Alain2017}.
Therefore, it is fair to expect that if the concept $\concept{C}$ being probed is indeed relevant for the task that a neural network $\main$ was trained for, then its internal representations have likely distilled some information that is related to $\concept{C}$. By leveraging these representations, one should be able to train a probing model $\probe$ to identify $\concept{C}$, even with few train data.

To assess this hypothesis, we train and test a probe $\probe$ of each considered probing architecture for each original model $\main$, while varying the amount of training data available. Figure \ref{fig:dataset_size} shows the average accuracy of each probing architecture over the considered layers for each model $\main$ for relevant and non-relevant concepts.
In the setting of the ImageNet and CUB datasets, the maximum amount of samples is fewer than in the others due to insufficient sample availability.

The results show that the accuracy of the probes grows quickly for relevant concepts, starting to stabilize at around $|\datas_\probe| = 200$ training samples.
Non-relevant concepts seem to require additional training samples while achieving significantly worse results.
We attribute these results to the fact that there are fewer incentives for a model to retain information in its inner representations regarding concepts that are not relevant to its task.
These results seem to support our conjecture, with the relevant concepts requiring fewer samples than the non-relevant ones. 

\begin{figure*}[t]
\centering
\includegraphics[width=\picscale\linewidth]{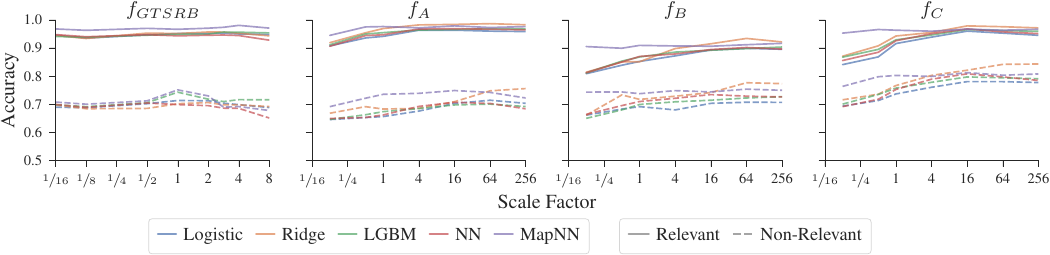}
\caption{Probe performance by scaling of the original model's size.}
\label{fig:representation_size}
\end{figure*}

We also observe that the probes using some kind of regularization -- such as the Ridge and MapNN -- seem to, on average, achieve better results. Furthermore, in the case of MapNN, a t-test shows that its accuracy is statistically significantly higher than the remaining probes when probing from $\nnimagenet$, $\nnfashion$, $\nncub$, $\nngtsrb$, $\nna$, $\nnb$, and $\nnc$, with the maximum amount of available data.
Additionally, Ridge seems to be the probe that is benefiting less from additional data, showing, on average, a smaller accuracy increase from $500$ to $1000$ samples than all other probe architectures.

These results indicate that one should be able to accurately probe relevant concepts with few training samples -- on average, the probe's performance reaches $97.3\%$ of its maximum measured value with as few as $250$ training samples (about $0.8\%$ of the original model's training data).
While the particular values may vary depending on the model being studied and concepts being probed, we consider these results to be encouraging, supporting that concept probing might still be feasible even when the amount of available data is limited.


\section{Effect of Original Model Size}
\label{sec:main_model_size}
\label{Sec:model_size}

Since concept probes are trained based on the activations of a neural network model, as models grow in size, so does the number of their activations, raising the question of how this impacts the training of probes. 
In this section, we investigate the relationship between the original model's size and the resulting probe's performance.

Consider two models $\main_1$ and $\main_2$, both trained to perform the same task and having achieved a similar test performance, where $\main_2$ is a scaled-up version of $\main_1$.
We hypothesize that if we probe both models using the same probe and data, the probing performance should remain unchanged or slightly deteriorate in $\main_2$, as the number of probed units increases without necessarily providing new information to help improve the probe's training, i.e., we expect the added units to mainly provide redundant information.

To test this hypothesis, we need to train multiple original models, scaling up or down their sizes while ensuring that the resulting model's accuracy was kept similar to the initial (without rescaling) original model.  
Given the time and computational resources required to train such scaled versions of the original models in settings where large models with pretraining were utilized, these experiments were only conducted in the settings of the XTRAINS and GTSRB datasets. 
For the models trained in the XTRAINS dataset, the convolutional part of the model was frozen, and only the dense layers were rescaled and retrained. 
A different approach was taken for the MobileNetV2 architecture, which was used in the setting of the GTSRB dataset, where there is no such separation. The models were retrained from scratch, rescaling the whole network architecture. 
Only models with a difference smaller than $1.5\%$ in test accuracy from the initial model were considered.

In Figure \ref{fig:representation_size}, we show the resulting average probe accuracy for the considered concepts in each dataset while varying the size of the original models. 
The explored scale factors differ for the GTSRB and XTRAINS datasets due to the large difference in the size of the models used in each setting.

Perhaps surprisingly, we observe that the accuracy of the probing models seems to slightly increase with the scaling up of the original models' size within the tested scale factors, which generate rather large models. The largest scale factors led, on average, to a $1.8\%$ test accuracy increase. This might be due to the additional features providing some new information that is beneficial for the probing models. 
Additionally, we observe that scaling down the models seems to deteriorate the probes' performance, especially for the smaller models used in the XTRAINS dataset, reducing the accuracy by $3.9\%$. This might indicate that, due to the limited size of the models' representations, the models had to somehow compress these representations, making the probing of each concept increasingly more complex.
A similar observation was made in \cite{Palsson2024}, where they probe from the most compressed layer of an autoencoder, and observe that the probing performance worsens, especially for linear probes. 
Interestingly, we observe that the mapping network probes seem mostly unaffected by the scaling down of the original model's size, likely because they are able to select units from various layers.

We conclude that, despite the vast number of features often considered by probing models, they are mostly able to successfully map them to their respective concepts. 
This is quite relevant because model sizes have been increasing in the last few years \cite{Villalobos2022}, indicating that concept probing is still feasible even when dealing with larger-sized models without compromising the probe's performance.


\section{Effect of Reusing Data}
\label{Sec:data_reuse}

\begin{figure*}[t]
\centering
\includegraphics[width=\picscale\linewidth]{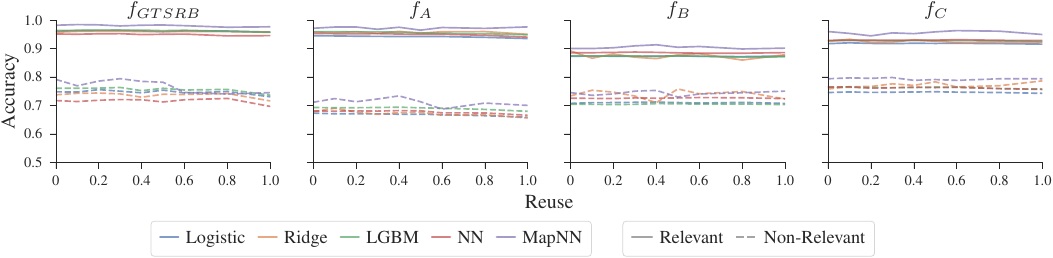}
\caption{Probe performance by percentage of data reuse from the original model's training data.}
\label{fig:data_reuse}
\end{figure*}

When training the probes, it is common practice to use the activations $\main_{u}(x^{(i)})$ generated by samples $x^{(i)}$ that were \emph{not} involved in the training of the original model $\main$, e.g., \cite{Palsson2024}.
This is usually done to avoid any undesirable effects that reusing the same data might have on the probe model's training. 
For example, one might hypothesize that the activation patterns of a neural network model $\main$ are different for samples used in $\main$'s training and for samples from the same domain that were not involved in its training.
If this were the case, one might expect that by training some probe $\probe$ on the activations produced by the original model's $\main$ training samples, the resulting probe might underperform when tested on samples that were not involved in $\main$'s training.
In this section, we examine the effect of reusing the same input samples $x^{(i)}$ from the original model's train set $\datas_\main$ on the probing model's train set $\datas_\probe$.

If reusing the input samples $x^{(i)}$ from the original model's train set $\datas_\main$ to train some probe model $\probe$ has some negative impact on the resulting probe's performance, then we should expect that the test performance of probe $\probe$ would decrease as the percentage of samples $x^{(i)}$ from the original model's train set $\datas_\main$ in the probe's training set $\datas_\probe$ increases. Otherwise, the resulting performance should remain relatively unchanged.

To test this effect, we need to train multiple probes while varying the amount of reused data from their respective original models' train set $\datas_\main$. 
Given that $\nnimagenet$, $\nnfashion$, and $\nncub$ were trained using all available training data and that $\nnplaces$'s training data is disjoint from its probe's training data, we could not reproduce this experiment in these settings, thus focusing on the GTSRB and XTRAINS datasets. 

In Figure \ref{fig:data_reuse}, we show the average performance of the trained probes when varying the amount of reused data. 
Interestingly, the performance of the trained probes $\probe$ remains mostly unchanged, even when all of $\probe$'s training data came from their respective original model's training set.
In fact, if we fit a line of best fit to the results, its slope is just slightly negative ($-4.1\!\times\!10^{-3}$), with a t-test showing that there is no statistically significant evidence that the amount of reuse affects performance, i.e., that this slope is different from zero.

While we did not observe any significant decrease in the model's performance when trained with the same data that was used to train its original model, this does not imply that there are no consequences for this decision.
Effects of reusing the original model's training data might manifest in other ways that are not measurable by a probe's test accuracy.
However, this provides us with a helpful insight: -- the resulting probe does not seem to overfit to the
original model's activations when trained with reused data, which is one of the main causes of concern on this matter.
These results are quite significant for domains where data might be scarce, and thus, being able to reuse the same data that was already gathered for the training of the original model might prove particularly useful.


\section{Effect of Data Quality}
\label{sec:data_quality}
\label{Sec:data_quality}

\begin{figure*}[t]
\centering
\includegraphics[width=\picscale\linewidth]{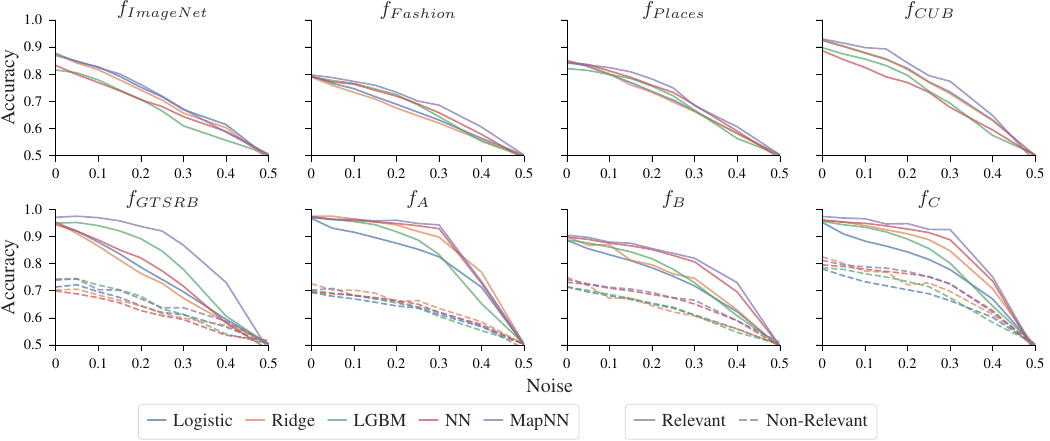}
\caption{Probe performance by amount of noise in data labels.}
\label{fig:data_quality}
\end{figure*}

Data quality is essential for any data-driven approach \cite{Schmarje2022}, such as machine learning. 
However, for a variety of reasons, data quality often falls short of being satisfactory \cite{Northcutt2021}.
Given the diagnostic nature of concept probing, where a model is trained to identify some concept of interest given the activations of another model, it is reasonable to inquire how data quality might impact a probe's performance.

There exist various ways to assess the quality of data \cite{Gong2023}. Here, we consider label accuracy, i.e., how accurate are the labels $c^{(i)}$ in a probe's dataset $\{\main_{u}(x^{(i)}), c^{(i)}\}$.
This seems relevant given that concept probing is generally applied on a different set of labels from the ones used to train the original model, which might not always be as carefully considered.
The CUB dataset illustrates this situation, where bird species constitute the main labels of the dataset, with the attribute labels being found to be noisy \cite{Zhao2019}.

To examine the effect of varying the quality of a probe's training data, we artificially introduced noise in the concept labels by inverting the label value of a random subset of the samples.
Figure \ref{fig:data_quality} shows the average probe accuracy for each original model while varying the percentage of noisy train samples. 

The probing models seem to be robust to a reasonable amount of noise. 
Introducing $20\%$ of noise in a probe's train labels led to a relative performance reduction of $9.3\%$ on average. 
The effect of the introduced noise seems to aggravate significantly when more than $30\%$ noise is added to the train data.

To further validate the effect of the artificially introduced noise and compare it with naturally occurring noise in the samples' labels, we train and test probes with the original attribute labels provided in the CUB dataset, which are known to be noisily labeled \cite{Zhao2019}.
For the probed concepts in the CUB dataset, we found that $18.2\%$ of the attribute labels were incorrectly specified, with the vast majority of these ($93.4\%$) being samples where a concept was mislabeled as being present.
Training with these labels led to a relative performance reduction of $13.1\%$, which is significantly higher than the one resulting from the artificially introduced noise. This is understandable since the noise in the original CUB labels is not random and mainly affects the positive samples.

While these results are encouraging -- suggesting that probes exhibit some robustness to noise in the labels -- they should encourage the community to validate the quality of the data used when applying concept probing, and also inspire and motivate, we hope, the development of high-quality benchmarking datasets designed with concept probing research in mind.


\section{Related Work}
\label{Sec:related_work}

Until recently, the most popular methods to interpret neural network models were saliency and attribution methods \cite{Sundararajan2017,Rebuffi2020,Ivanovs2021}, which explain a model's prediction by providing a set of input features and their corresponding contributions, and proxy-based methods \cite{Schmitz1999,Augasta2012,Ribeiro2016}, which generate an interpretable model with similar behavior to the original one.
However, user studies \cite{Adebayo2020,Chu2020,Shen2020} showed that the explanations provided by these methods were often disregarded or unhelpful to end users.
This might be attributed to these methods explaining a model's behavior in terms of its input features, which might not always be meaningful or understandable to users. For example, highlighting a set of pixels from an image to a user does not necessarily convey to the user the meaning of such pixels or how they were used by the model.  
All of this guesswork still has to be done by the user.

Concept-based explainable AI \cite{Poeta2023,HeeLee2023,Schwalbe2024,HeeLee2024} emerged from the human necessity to interpret models through concepts (symbols) that are relevant and understandable to them, abstracting away from the particular input features of a model.
Multiple works have been based on the idea of relating the internal representations of neural network models to human-defined concepts of interest. Some attempt to interpret individual units of a model \cite{Mu2020}, while others attempt to interpret a layer \cite{Kim2018,Crabbe2022}, or even whole models \cite{Horta2021}.

Concept probing also stemmed from such ideas and became one of the most prevalent frameworks for interpreting neural network models in a concept-based manner.
Numerous works have been developed based on concept probing, some focus on examining what it tells us about the model being probed \cite{Alain2017,Pimentel2020}, while others focus on the architectures of concept probing models \cite{Pimentel2020a,Sanh2021,Zhou2021}, and others investigate concept probing on different kinds of models, such as LSTMs \cite{Linzen2016} and RNNs \cite{Hupkes2018}, or in different domains, like natural language processing \cite{Tenney2019} and game playing \cite{Palsson2024}.
Others have leveraged concept probing to develop methods to interpret or explain neural networks, e.g., to induce theories describing a model's internal classification process \cite{Ferreira2022}, to test whether a model's internal representations are consistent with some logic theory \cite{Lovering2022}, or to produce symbolic justifications for a model's outputs \cite{deSousaRibeiro2021}.

Despite the growing body of work on concept probing, there is a lack of work on understanding how the data used to train probing models affects the probes' results, as discussed in \cite{Belinkov2022}. 
This gap in the literature is significant, as concept probes are often used to assess properties regarding the probed model, and their development is underpinned by their training data. 
While there exists some work where various datasets and models are probed \cite{deSousaRibeiro2025NESY,Gurnee2024,Belinkov2017}, these do not address the effects that the data has on the probing models, focusing instead on what can be inferred regarding the probed models.

As mentioned in Section \ref{sec:main_model_size}, the mapping network probes, which use a procedure to select the units from which they probe, kept their probing accuracy even when probing from models with a smaller size.
This corroborates the fact that the units used to probe a concept highly influence the resulting probe's accuracy. \cite{Durrani2020} describes the use of elastic-net regularization to identify which units in a layer encode a given concept.

Throughout this work, we have shown that the results of probing for concepts relevant to a given model's task are substantially different from those of concepts that are not relevant.
While the topic of how to identify relevant concepts is outside the scope of this paper, we point out that there exists research aimed at discovering concepts that are relevant for a model's predictions \cite{Horta2021,Ghorbani2019}, which might help practitioners identify concepts to probe for.


\section{Conclusions}
\label{Sec:conclusions}

In this paper, we investigated the effect that the data used to train concept probes has on their resulting performance on image classification tasks.
To this end, we performed a thorough experimental evaluation with datasets with varied characteristics, using different kinds of probes on neural network models with diverse architectures.

We conclude that concept probing generally requires few train data for concepts that are relevant to the probed model's task, even -- perhaps surprisingly -- when probing the representations of large models, which provides additional evidence that probing models are capable of leveraging the knowledge distilled within the probed model's representations to establish a mapping to their respective concepts.
This evidence is further strengthened by the observation that probing models for non-relevant concepts achieve a significantly worse test performance while requiring additional training data.

We also found that repurposing the data involved in the training of the probed model to train probing models does not degrade their resulting test accuracy, providing a valuable insight which suggests that this might be a viable approach in settings where data is limited.

We further conclude that, despite the probing models' moderate robustness when faced with low-quality data in the form of mislabeled train samples, more effort should be put into assessing the quality of data. 
This is particularly relevant given the kind of inferences about models that are made based on concept probing, and given that well-known machine learning datasets often have low quality \cite{Northcutt2021}.

Our experiments can also be used to compare the behavior of the most popular probing architectures on various image classification neural network models and datasets.

To the best of our knowledge, this study is the first to 
examine the effect that such properties have on the development of concept probing models, providing a starting point towards more research on how data affects concept probing, a topic that has been largely neglected \cite{Belinkov2022}.
We hope this work strengthens confidence in the reproducibility and reliability of concept probing, while providing evidence on how it is influenced by the quantity, provenance, and quality of the data, as well as the size of the probed model -- helping to counter the spread of unverified folk theories about its applicability.


\begin{ack}
This work was supported by FCT I.P. through UID/04516/NOVA Laboratory for Computer
Science and Informatics (NOVA LINCS) and through PhD grant (DOI 10.54499/UI/BD/
151266/2021), and by Project Sustainable Stone by Portugal - Valorization of Natural Stone for a digital, sustainable and qualified future, no 40, proposal C644943391-00000051, co-financed by PRR - Recovery and Resilience Plan of the European Union (Next Generation EU).
\end{ack}


\bibliography{biblio}

\clearpage
\appendix

\renewcommand{\thefigure}{A.\arabic{figure}}
\renewcommand{\thetable}{A.\arabic{table}}
\setcounter{figure}{0}
\setcounter{table}{0}


\section{Additional Caltech-UCSD Birds-200-2011 Labels} \label{Sec:appendix_cub}

In this Appendix, we describe the procedure used to relabel a subset of the Caltech-UCSD Birds-200-2011 dataset (CUB) \cite{Wah2011} attributes.

As mentioned in \cite{Zhao2019, Koh2020} some of the attributes in this dataset were found to be noisily labeled, which prompted us to perform a relabeling of the attribute labels involved in this work in order to guarantee the quality of the data used throughout the experiments.

From the $312$ attributes in this dataset, a random subset of $11$ attributes was considered: $\sf \exists has\-Bill\-Shape.Nee\-dle$, $\sf \exists has\-Bill\-Shape.Hooked\-Sea\-bird$, $\sf \exists has\-Head\-Pat\-tern.Striped$, $\sf \exists has\-Breast\-Col\-or.Yel\-low$, $\sf \exists has\-Throat\-Col\-or.Red$, $\sf \exists has\-Eye\-Col\-or.Red$, $\sf \exists has\-Bel\-ly\-Col\-or.Blue$, $\sf \exists has\-Bel\-ly\-Col\-or.Yel\-low$, $\sf \exists has\-Shape.Duck\-Li\-ke$, $\sf \exists has\-Crown\-Col\-or.White$, $\sf \exists has\-Crown\-Col\-or.Red$.

These labels are made available in \citep{cub_relabel_2025}.

\subsection{Relabeling Procedure}

For each of the $11$ selected attributes, the following procedure was performed: 

\begin{enumerate}
	\item Split the data into train and test images, according to the splits defined in the dataset.
	\item For each of the train and test sets:
	\begin{enumerate}
		\item Split the data into a positive set ($P$), where the attribute is present, and a negative set ($N$), where the attribute is absent, of samples, according to the original labels in the dataset.
		\item Define a constant $n = min(\{|P|,\ |N|,\ 500\})$
		\item Randomly select $n$ samples from $P$:
		\begin{enumerate}
			\item Relabel the selected samples.
		\end{enumerate}
		\item Randomly select $n$ samples from $N$:
		\begin{enumerate}
			\item Relabel the selected samples.
		\end{enumerate}
	\end{enumerate}
\end{enumerate}

This procedure produces a new set of labels for a subset of samples of the original dataset, for a given attribute.

The relabeling of each image was performed according to the descriptions provided in the field guide \cite{whatbird} referred in \cite{Wah2011}. The relabeling of the attributes was made with respect to whether the attributes were visible in each image. Thus, even if a bird of a given species is known to have a certain attribute, but it is not visible in the image, the attribute is labeled as being absent from the image.

\subsection{Label Statistics}

The distribution of positive and negative samples in the original labels is shown in Table \ref{tab:stats1}. Table \ref{tab:stats2} shows the resulting number of samples for each attribute after executing the described procedure. Table \ref{tab:stats3} summarizes the identified discrepancies -- instances originally labeled as positive but found to be negative, and instances labeled as negative but found to be positive.

\begin{table}[tb]
	{\center
		{\scriptsize
	\begin{tabular}{@{}rcc@{}}
		\toprule
		Attribute & \# Original Positive & \# Original Negative \\ \midrule
		$\mathsf{\exists hasBillShape.Needle}$         & 289                  & 11499                \\
		$\mathsf{\exists hasBillShape.HookedSeabird}$         & 751                  & 11037                 \\
		$\mathsf{\exists hasHeadPattern.Striped}$         & 805                  & 10983                \\
		$\mathsf{\exists hasBreastColor.Yellow}$         & 1613                 & 10175                  \\
		$\mathsf{\exists hasThroatColor.Red}$         & 453                  & 11335                \\
		$\mathsf{\exists hasEyeColor.Red}$         & 278                  & 11510                 \\
		$\mathsf{\exists hasBellyColor.Blue}$         & 271                  & 11517                \\
		$\mathsf{\exists hasBellyColor.Yellow}$         & 1692                 & 10096               \\
		$\mathsf{\exists hasShape.DuckLike}$         & 723                  & 11065                \\
		$\mathsf{\exists hasCrownColor.White}$         & 1433                 & 10355               \\
		$\mathsf{\exists hasCrownColor.Red}$         & 572                  & 11216               \\ 
		\bottomrule
	\end{tabular}
	}
	\caption{Number of samples for each attribute in the original dataset.}
	\label{tab:stats1}
	}
\end{table}

\begin{table}[tb]
	{\center
		{\scriptsize
			\begin{tabular}{@{}rcc@{}}
				\toprule
				Attribute & \# Relabeled Positive & \# Relabeled Negative \\ \midrule
				$\mathsf{\exists hasBillShape.Needle}$         &  190                   & 414                   \\
				$\mathsf{\exists hasBillShape.HookedSeabird}$         &  629                   & 875                    \\
				$\mathsf{\exists hasHeadPattern.Striped}$         &  324                   & 675                   \\
				$\mathsf{\exists hasBreastColor.Yellow}$         &  387                   & 642                   \\
				$\mathsf{\exists hasThroatColor.Red}$         &  304                   & 678                   \\
				$\mathsf{\exists hasEyeColor.Red}$         &  208                   & 382                   \\
				$\mathsf{\exists hasBellyColor.Blue}$         &  141                   & 436                   \\
				$\mathsf{\exists hasBellyColor.Yellow}$         &  371                   & 639                   \\
				$\mathsf{\exists hasShape.DuckLike}$         &  502                   & 922                    \\
				$\mathsf{\exists hasCrownColor.White}$         &  245                   & 760                   \\
				$\mathsf{\exists hasCrownColor.Red}$         &  401                   & 785                     \\ \bottomrule
			\end{tabular}
		}
		\caption{Number of samples for each attribute after executing the relabeling procedure.}
		\label{tab:stats2}
	}
\end{table}

\begin{table}[tb]
	{\center
		{\scriptsize
			\begin{tabular}{@{}rcc@{}}
				\toprule
				Attribute & \# Mislabeled Positive & \# Mislabeled Negative \\ \midrule
				$\mathsf{\exists hasBillShape.Needle}$         &  99                     & 0                      \\
				$\mathsf{\exists hasBillShape.HookedSeabird}$         &  153                    & 31                     \\
				$\mathsf{\exists hasHeadPattern.Striped}$         &  234                    & 59                     \\
				$\mathsf{\exists hasBreastColor.Yellow}$         &  137                    & 6                      \\
				$\mathsf{\exists hasThroatColor.Red}$         &  151                    & 2                      \\
				$\mathsf{\exists hasEyeColor.Red}$         &  78                     & 8                      \\
				$\mathsf{\exists hasBellyColor.Blue}$         &  130                    & 0                      \\
				$\mathsf{\exists hasBellyColor.Yellow}$         &  151                    & 15                     \\
				$\mathsf{\exists hasShape.DuckLike}$         &  239                    & 18                     \\
				$\mathsf{\exists hasCrownColor.White}$         &  259                    & 4                      \\
				$\mathsf{\exists hasCrownColor.Red}$         &  177                    & 6                      \\ \bottomrule
			\end{tabular}
		}
		\caption{Number of instances originally labeled as positive but found to be negative (\# Mislabel Positive), and instances labeled as negative but found to be positive (\# Mislabeled Negative), for each attribute.}
		\label{tab:stats3}
	}
\end{table}

\subsection{Examples of Attribute Labeling Discrepancies}

\begin{figure*}
	\centering
	\includegraphics[scale=0.80]{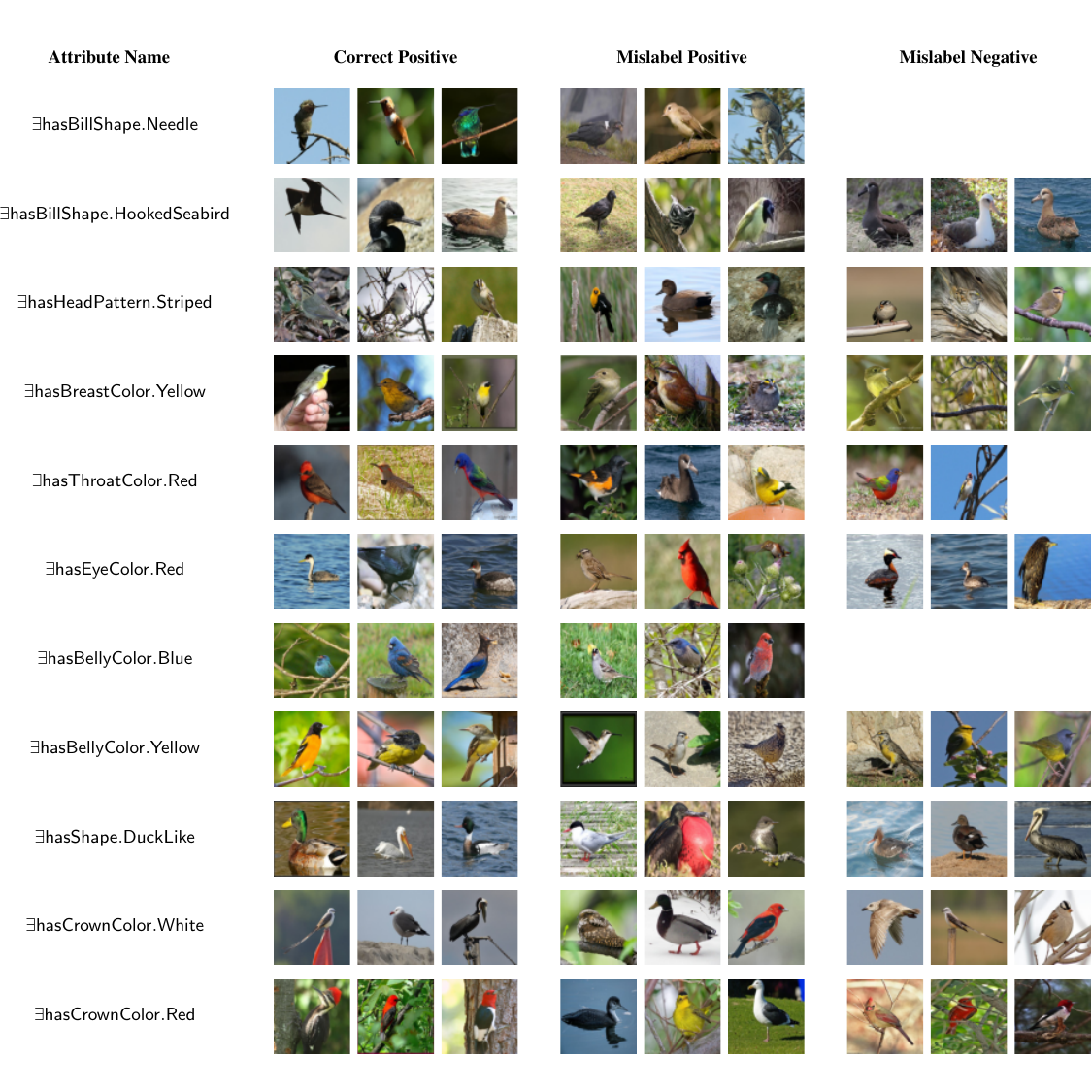}
	\caption{Instances of correctly labeled samples and mislabeled samples.}
	\label{fig:cub_data_quality}
\end{figure*}

Figure \ref{fig:cub_data_quality} shows instances for each attribute, illustrating three cases: i) correctly labeled samples where the attribute is present; ii) mislabeled samples where the attribute is absent but originally labeled as present; iii) and mislabeled samples where the attribute is present but originally labeled as absent.

\clearpage

\renewcommand{\thefigure}{B.\arabic{figure}}
\renewcommand{\thetable}{B.\arabic{table}}
\setcounter{figure}{0}
\setcounter{table}{0}


\section{Additional German Traffic Sign Recognition Benchmark  Labels and Ontology}
\label{Sec:appendix_gtsrb}

In this Appendix, we describe the procedure used to label a subset of the German Traffic Sign Recognition Benchmark  dataset (GTSRB) \cite{Stallkamp2011} regarding additional concepts of interest defined in the 1968 Convention on Road Signs and Signals \cite{eutrafficconvention}, which describes each type of traffic sign in the dataset based on visual features.

These labels and ontology are made available in \citep{gtsrb_concepts_2025}.

\subsection{Ontology Development}

This ontology was obtained by translating to Description Logics all of the excerpts related to the $43$ traffic signs included in the GTSRB dataset from the 1968 Convention on Road Signs and Signals \cite{eutrafficconvention}.

The ontology was designed by identifying concepts and roles, and forming axioms that reflect as close as possible the descriptions provided in \cite{eutrafficconvention} regarding the traffic signs considered in the GTSRB dataset. Additionally, axioms reflecting the hierarchies and categories of the traffic signs in the GTSRB dataset, were defined based on the descriptions in \cite{eutrafficconvention}.

This process resulted in an ontology that describes the $43$ signs existing in the GTSRB dataset, and how they relate to the $4$ traffic signal categories, based on the following vocabulary:
\begin{itemize}
\item Shape -- a sign may have a triangular, circular, rectangular, octagonal, or diamond shape;
\item Ground -- a sign may have a white, yellow, blue, red, or orange ground;
\item Border -- a sign may have a white, red, or black border;
\item Bar -- a sign may have a white or black bar;
\item Symbol -- a sign may have symbols; $35$ different symbols were considered.
\end{itemize}

Three different kinds of information were provided in \cite{eutrafficconvention} regarding the traffic signals and their hierarchy:
\begin{itemize}
\item Information describing the features of a given traffic signal, e.g., the following excerpt provides information regarding the features of a `Model B, 2a Stop sign': \\
``Model B, 2a is octagonal with a red ground bearing the  word "STOP" in white in English or in the  language of the State concerned;'', which resulted in the axiom:\\ $ \sf \exists hasShape.Octagon \sqcap \exists hasGround.Red \sqcap \exists hasSymbol.(White \sqcap Stop) \equiv B2a$.

\item Information describing which traffic signals belong to a given traffic signal class, e.g., the following excerpt provides information regarding which traffic signs belong to the class `A.1. Danger Warning signs': \\
``Section A DANGER WARNING SIGNS (...) 1. Dangerous bend or bends: \\
Warning of a dangerous bend or succession of dangerous bends shall be given by one of the following symbols, whichever is appropriate:
	\begin{enumerate}[label=(\alph*)]
	\item A, 1\textsuperscript{a}: left bend
	\item A, 1\textsuperscript{b}: right bend
	\item A, 1\textsuperscript{c}: double bend, or succession of more than two bends, the first to the left
	\item A, 1\textsuperscript{d}: double bend, or succession of more than two bends, the first to the right.''
	\end{enumerate}
	which resulted in the following axiom: $\sf A1a \sqcup A1b \sqcup A1c \sqcup A1d \sqsubseteq A1$;

\item Information describing how different classes of traffic signals are categorized, e.g., the following excerpt provides information regarding which traffic signs classes belong to the category `A. Danger warning signs': \\
``The ``A'' DANGER WARNING signs shall be of model A\textsuperscript{a} or model A\textsuperscript{b} both described here and reproduced in Annex 3, except signs A, 28 and A, 29 described in paragraphs 28 and 29 below respectively.'', which resulted in the axiom: $\sf A^a \sqcup A^b \sqcup A28 \sqcup A29 \equiv A$;

\end{itemize}

To ensure that the ontology properly describes all traffic signals existing in the GTSRB dataset, we added the descriptions of traffic signs that were introduced after the 1968 Convention on Road Signs and Signals, such as the signs: A 33, A 34, and C 3\textsuperscript{e3}.

The concept $\sf Pole$, representing a traffic signal's pole was also added to the ontology. This non-relevant concept is not involved in any of the ontology's axioms, or even directly involved in the task performed by the main network $\nngtsrb$. 

\begin{figure*}
	 \centering \hfill \hfill 
	\includegraphics[height=0.95\textheight]{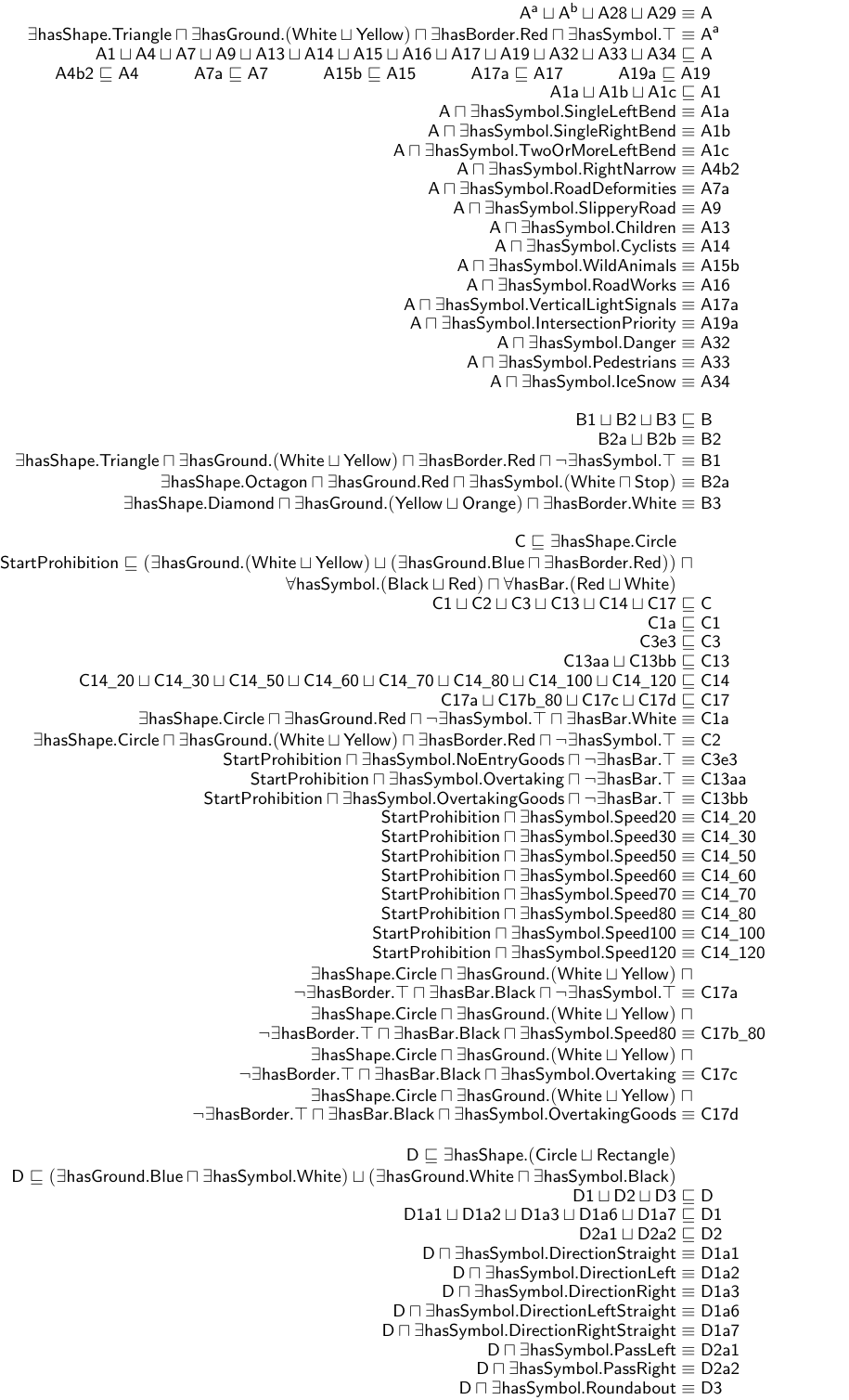} \hfill
	\caption{Resulting ontology describing the GTSRB domain.}
	\label{fig:ontology}
\end{figure*}

The complete ontology can be found in Figure \ref{fig:ontology}.

\subsection{Labeling Procedure}

From the resulting ontology the following $41$ of interest were considered for labeling: $\sf \exists hasBar.\top$, $\sf \exists hasBar.Black$, $\sf \exists hasBar.White$, $\sf \exists hasBorder.\allowbreak\top$, $\sf \exists hasBorder.\allowbreak Black$, $\sf \exists hasBorder.Red$, $\sf \exists hasBorder.\allowbreak White$, $\sf \exists hasGround.Red$, $\sf \exists hasGround.(White \sqcup Yellow)$, $\sf \exists  has \allowbreak Ground.\allowbreak Yellow$, $\sf Start\- Prohibition$, $\sf End\- Prohibition$, $\sf \exists hasShape.Circle$, $\sf \exists has\- Shape.\allowbreak Diamond$, $\sf \exists hasShape.Triangle$, $\sf \exists hasShape.Octagon$, $\sf \exists has\- Symbol.\allowbreak \top$, $\sf \exists has\- Symbol.\allowbreak Black$, $\sf \exists has\- Symbol.\allowbreak White$, $\sf \exists has\- Symbol.\allowbreak NoEntryGoods$, $\sf \exists has\- Symbol.\allowbreak Overtaking$, $\sf \exists has\- Symbol.\allowbreak OvertakingGoods$, $\sf \exists has\- Symbol.\allowbreak Speed20$, $\sf \exists has\- Symbol.\allowbreak Speed30$, $\sf \exists has\- Symbol.\allowbreak Speed50$, $\sf \exists has\- Symbol.\allowbreak Speed60$, $\sf \exists has\- Symbol.\allowbreak Speed70$, $\sf \exists has\- Symbol.\allowbreak Speed80$, $\sf \exists has\- Symbol.\allowbreak Speed100$, $\sf \exists has\- Symbol.\allowbreak Speed120$, $\sf \exists has\- Symbol.\allowbreak Stop$, $\sf D1a1$, $\sf D1a4$, $\sf D1a5$, $\sf D1a6$, $\sf D1a7$, $\sf D2a1$, $\sf D2a2$, $\sf D3$, $\sf Blue$, $\sf Pole$.

For each of these concepts, a random subset of $200$ samples from each of the $43$ traffic sign classes contained in the training set of GTSRB was manually labeled. Similarly, for testing purposes, a random subset of $60$ images from each of the $43$ traffic sign classes was labeled.

The resulting distribution of positive and negative samples in the train and test sets of the GTSRB dataset is shown in Table \ref{tab:gtsrb_stats}.


\begin{table*}[tb]
	\centering
	\begin{tabular}{@{}rcccc@{}}
		\toprule
		\multicolumn{1}{c}{\multirow{2}{*}{Concept}}  & \multicolumn{2}{c}{Train}                 & \multicolumn{2}{c}{Test}                  \\
		\multicolumn{1}{c}{}                          & \# Positive Samples & \# Negative Samples & \# Positive Samples & \# Negative Samples \\ \midrule
		$\sf \exists hasBar.\top$                     & 1178                & 7422                & 350                 & 2230                \\
		$\sf \exists hasBar.Black$                    & 870                 & 7730                & 251                 & 2329                \\
		$\sf \exists hasBar.White$                    & 309                 & 8291                & 92                  & 2488                \\
		$\sf \exists hasBorder.\top$                  & 5832                & 2768                & 1761                & 819                 \\
		$\sf \exists hasBorder.Black$                 & 1018                & 7582                & 359                 & 2221                \\
		$\sf \exists hasBorder.Red$                   & 4710                & 3890                & 1367                & 1213                \\
		$\sf \exists hasBorder.White$                 & 202                 & 8398                & 57                  & 2523                \\
		$\sf \exists hasGround.Red$                   & 377                 & 8223                & 109                 & 2471                \\
		$\sf \exists hasGround.(White \sqcup Yellow)$ & 6451                & 2149                & 1941                & 639                 \\
		$\sf \exists hasGround.Yellow$                & 262                 & 8338                & 72                  & 2508                \\
		$\sf StartProhibition$                        & 2660                & 5940                & 795                 & 1785                \\
		$\sf EndProhibition$                          & 810                 & 7790                & 240                 & 2340                \\
		$\sf \exists hasShape.Circle$                 & 5058                & 3542                & 1524                & 1056                \\
		$\sf \exists hasShape.Diamond$                & 202                 & 8398                & 60                  & 2520                \\
		$\sf \exists hasShape.Triangle$               & 3313                & 5287                & 999                 & 1581                \\
		$\sf \exists hasShape.Octagon$                & 200                 & 8400                & 60                  & 2520                \\
		$\sf \exists hasSymbol.\top$                  & 7612                & 988                 & 2290                & 290                 \\
		$\sf \exists hasSymbol.Black$                 & 5611                & 2989                & 1696                & 884                 \\
		$\sf \exists hasSymbol.White$                 & 1820                & 6780                & 544                 & 2036                \\
		$\sf \exists hasSymbol.NoEntryGoods$          & 166                 & 8434                & 51                  & 2529                \\
		$\sf \exists hasSymbol.Overtaking$            & 306                 & 8294                & 97                  & 2483                \\
		$\sf \exists hasSymbol.OvertakingGoods$       & 310                 & 8290                & 86                  & 2494                \\
		$\sf \exists hasSymbol.Speed20$               & 196                 & 8404                & 60                  & 2520                \\
		$\sf \exists hasSymbol.Speed30$               & 186                 & 8414                & 59                  & 2521                \\
		$\sf \exists hasSymbol.Speed50$               & 178                 & 8422                & 54                  & 2526                \\
		$\sf \exists hasSymbol.Speed60$               & 180                 & 8420                & 57                  & 2523                \\
		$\sf \exists hasSymbol.Speed70$               & 188                 & 8412                & 60                  & 2520                \\
		$\sf \exists hasSymbol.Speed80$               & 348                 & 8252                & 102                 & 2478                \\
		$\sf \exists hasSymbol.Speed100$              & 164                 & 8436                & 58                  & 2522                \\
		$\sf \exists hasSymbol.Speed120$              & 146                 & 8454                & 43                  & 2537                \\
		$\sf \exists hasSymbol.Stop$                  & 196                 & 8404                & 58                  & 2522                \\
		$\sf D1a1$                                    & 200                 & 8400                & 60                  & 2520                \\
		$\sf D1a4$                                    & 200                 & 8400                & 60                  & 2520                \\
		$\sf D1a5$                                    & 200                 & 8400                & 60                  & 2520                \\
		$\sf D1a6$                                    & 200                 & 8400                & 60                  & 2520                \\
		$\sf D1a7$                                    & 200                 & 8400                & 60                  & 2520                \\
		$\sf D2a1$                                    & 200                 & 8400                & 60                  & 2520                \\
		$\sf D2a2$                                    & 200                 & 8400                & 60                  & 2520                \\
		$\sf D3$                                      & 200                 & 8400                & 60                  & 2520                \\
		$\sf Blue$                                    & 2138                & 6462                & 686                 & 1894                \\
		$\sf Pole$                                    & 3413                & 5187                & 863                 & 1717                \\ \bottomrule
	\end{tabular}
	\caption{Number of samples for each labeled concept of interest in the dataset.}
	\label{tab:gtsrb_stats}
\end{table*}

\begin{figure*}
	\centering
	\includegraphics[scale=0.90]{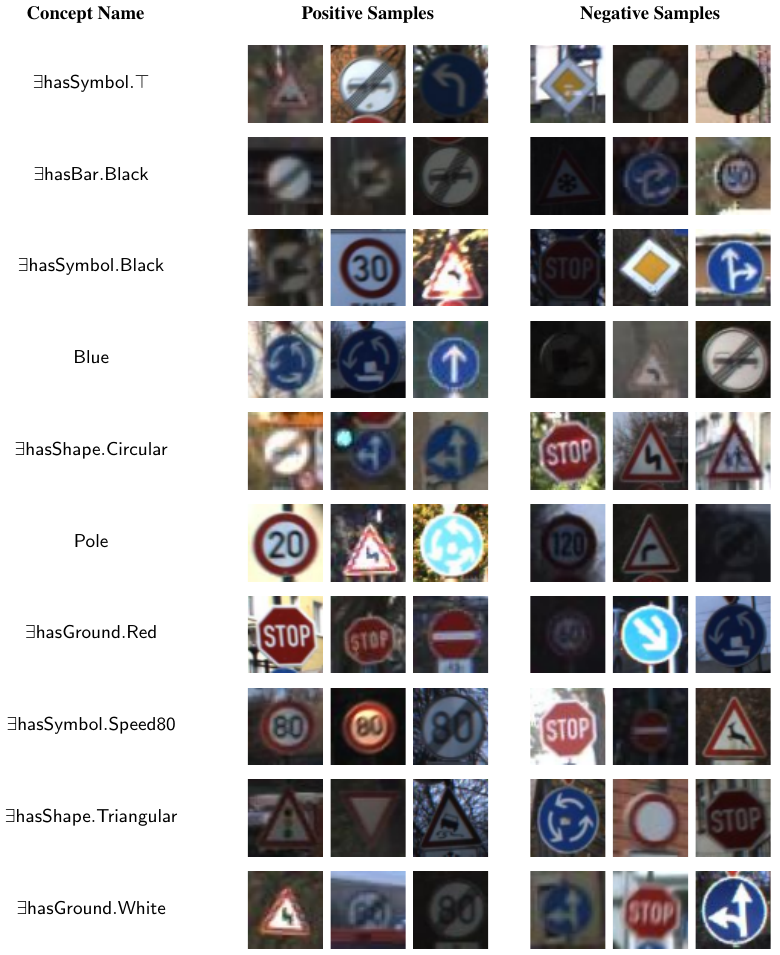}
	\caption{Positive and negative samples for each concept.}
	\label{fig:gtsrb_sample_images}
\end{figure*}

Figure \ref{fig:gtsrb_sample_images} shows random instances of samples where the considered concepts was present (positive samples) and absent (negative samples).

\end{document}